\documentclass{article}

\usepackage[preprint]{neurips_2026}

\usepackage[T1]{fontenc}
\usepackage[utf8]{inputenc}
\usepackage{amsmath,amsfonts}
\usepackage{graphicx}
\usepackage{booktabs}
\usepackage{placeins}
\usepackage{multirow}
\usepackage{multirow}
\usepackage{xcolor} 
\usepackage{graphicx}
\usepackage{bbm}
\usepackage[pagebackref,breaklinks,colorlinks]{hyperref}
\usepackage[table]{xcolor}
\definecolor{softredbg}{RGB}{255,220,220}
\newcommand{\best}[1]{\cellcolor{softredbg}\textbf{#1}}
\title{Spackle: Completing Large View Single Image NVS with Adaptive Gaussians}

\author{%
  \textbf{Xuanzhi Liu}\textsuperscript{1} \quad
  \textbf{Yuhe Zhou}\textsuperscript{1} \quad
  \textbf{Xinyi Wu}\textsuperscript{2} \quad
  \textbf{Zhenyao Wu}\textsuperscript{2}
  \\[3pt]
  \textbf{Jinghao Chen}\textsuperscript{2} \quad
  \textbf{Ruize Han}\textsuperscript{1} \quad
  \textbf{Song Wang}\textsuperscript{1}
  \\[8pt]
  {\normalsize \textsuperscript{1}Shenzhen University of Advanced Technology}
  \\[3pt]
  {\normalsize \textsuperscript{2}HONOR}
}
\begin{document}

\maketitle

\begin{abstract}
Single-image novel view synthesis (NVS) enables photorealistic rendering of unobserved viewpoints from a single input. Practical NVS systems require two key capabilities: robust reconstruction of occluded regions and high inference efficiency. While hybrid decoupled frameworks combining feedforward 3D Gaussian Splatting (3DGS) and diffusion models show promise for large-view-deviation NVS, they suffer from capacity competition: a fixed number of Gaussians forces resource shifts from visible to newly disoccluded areas, degrading original scene fidelity when the target view deviates significantly from the input. To address this, we propose Spackle, a lightweight residual learning framework that mitigates capacity competition without sacrificing efficiency. Spackle operates in three stages: predicting base 3DGS attributes from given views, automatically identifying poorly reconstructed regions, and learning a residual 3DGS optimized exclusively for these areas. 
%Spackle automatically identifying poorly reconstructed regions from pretrained 3DGS models and learning a small residual 3DGS optimized exclusively for these areas.
At inference, we combine the baseline and augmented Gaussians for NVS. We conduct comprehensive experiments and show that Spackle achieves state-of-the-art performance on large-view-deviation cases.
\end{abstract}

\section{Introduction}

% Xuanzhi, take a look at these to see whether they reflect what you want to say. -- Song
Single-image novel view synthesis (NVS) seeks to generate photorealistic renderings of a scene from unobserved viewpoints using only a single input image. This technology serves as a critical enabler for high-impact applications spanning immersive virtual/augmented reality systems~\cite{broxton2020immersive}, embodied AI agents requiring spatial awareness~\cite{chhablani2025embodiedsplat}, and automated 3D content creation pipelines~\cite{tang2023dreamgaussian}. Two key considerations guide the development of a practical single-image novel view synthesis (NVS) model: 1) robustly reconstructing both geometric structure and appearance details for occluded or unseen regions, and 2) delivering inference efficiency suitable for real-world latency-sensitive tasks. This paper aims to address both objectives simultaneously by proposing a novel framework capable of synthesizing high-fidelity novel views efficiently even when \emph{the target viewpoint deviates significantly from the input camera pose}—a long-standing challenge in the field. 

Our work draws inspiration from recent hybrid NVS frameworks that combine two complementary paradigms: 1) feedforward models predicting renderable 3D Gaussian Splatting (3DGS) scene representations in a single forward pass for extreme inference efficiency, and 2) large-scale pretrained diffusion models (e.g., Stable Diffusion) capable of generating photorealistic content for disoccluded region inpainting~\cite{ren2025gen3c,yu2024viewcrafter,duan2026navcrafter,cao2025mvgenmaster,kong2025causnvs}. In particular, as suggested in ~\cite{bahmani2025lyra,li2025flashworld,li2026syncfix,wu2025difix3d+,shen2026lyra2}, we first leverage computationally expensive diffusion models to synthesize large-view-change image pairs, which are then used to train efficient feedforward 3DGS-based NVS models. This decoupling strategy preserves real-time inference efficiency by eliminating reliance on generative model sampling during deployment, while still enabling generalization to large viewpoint deviations via high-quality synthetic training data~\cite{bahmani2025lyra,li2025flashworld,li2026syncfix,wu2025difix3d+,shen2026lyra2}.

\begin{figure}[!t]
    \centering
    \includegraphics[
        width=\linewidth,
        height=0.8\textheight,
        keepaspectratio
    ]{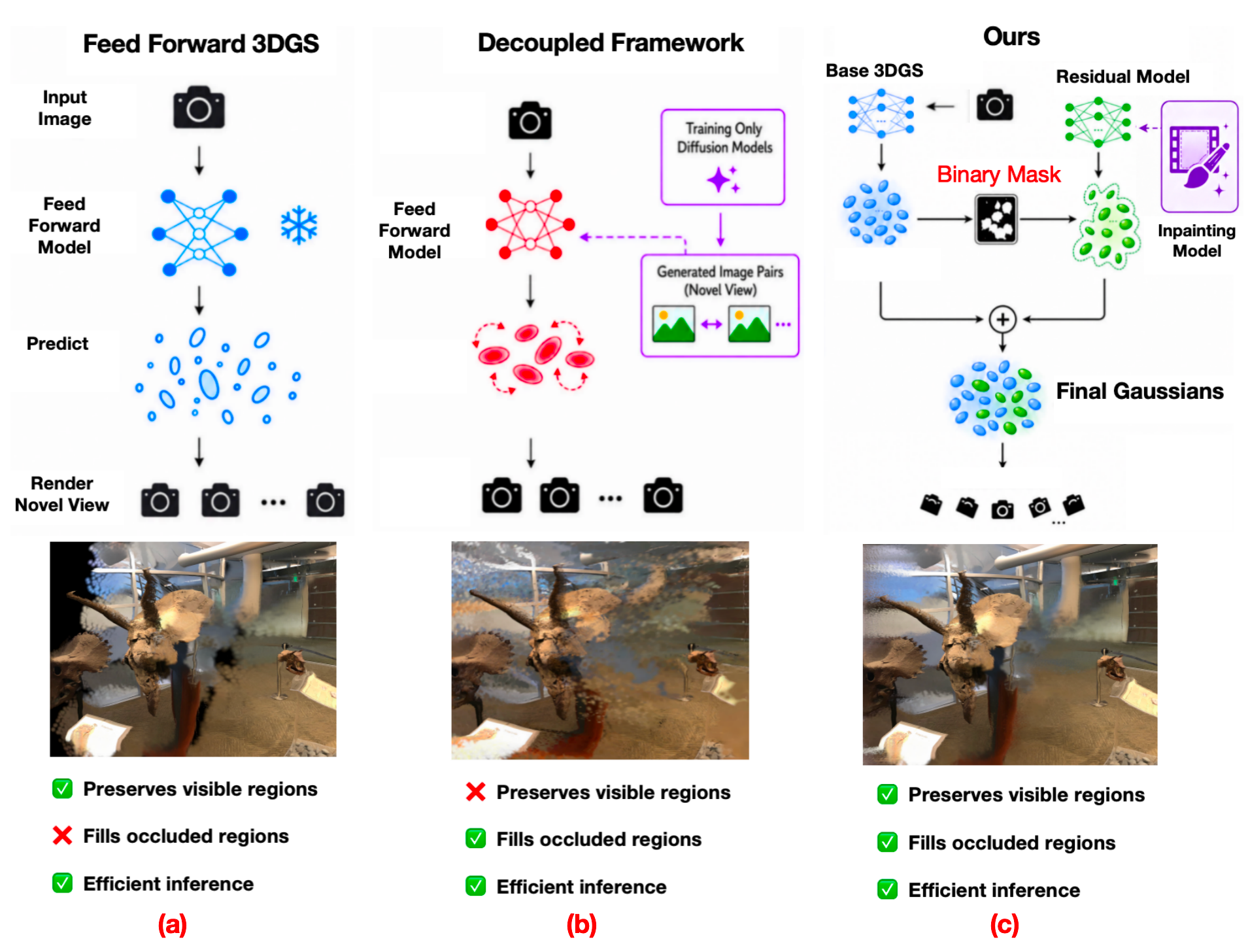}
    \vspace{-10pt}
\caption{
%Comparison of single-image NVS paradigms. (a) Feed-forward 3DGS is efficient but suffers from unfilled disoccluded regions. (b) Generative models can produce photorealistic novel views, yet with extremely high inference cost. (c) The baseline decoupled framework, which uses generated image pairs to train feed-forward 3DGS, inpaints disoccluded regions but exhibits capacity competition. (d) Our residual-learning framework mitigates capacity competition while preserving computational efficiency.
Comparison of single-image NVS paradigms. (a) Feed-forward 3DGS is efficient and preserves visible regions but leaves disoccluded regions unfilled. (b) Decoupled framework inpaints disoccluded regions but suffers from capacity competition. (c) Our residual-learning framework mitigates capacity competition, effectively fills disoccluded regions, and preserves visible regions while maintaining computational efficiency.
}
\label{fig:patchgs_overview}
\end{figure}

However, we observe that naive implementation of this decoupled framework introduces an unexpected failure mode: when synthesizing content for disoccluded regions, the model inadvertently distorts or alters the geometry and appearance of previously visible (seen) regions, leading to significant performance degradation in single-image NVS (Fig.~\ref{fig:patchgs_overview}). We attribute this issue to a novel phenomenon we term \emph{capacity competition}: existing methods operate with a fixed, predetermined number of 3D Gaussians during both training and inference. This fixed allocation proves insufficient when handling large viewpoint deviations, which require modeling extensive disoccluded regions. In such cases, the model reallocates a disproportionate number of Gaussians originally assigned to represent seen regions to newly visible areas, resulting in degraded fidelity and detail loss in the original viewpoint's content. 

% Recent progress suggests two complementary paradigms for addressing this problem. 
% The first paradigm is feed forward 3D reconstruction, where models directly predict geometry or explicit renderable representations from image observations in a single forward pass~\cite{mescheder2025sharp,wang2025vggt,szymanowicz2025bolt3d,szymanowicz2025flash3d,wang2026one2scene,rahary2026oneview}. 
% In particular, feed forward 3D Gaussian Splatting methods provide explicit Gaussian scene representations and support efficient, real time rendering without per scene optimization~\cite{mescheder2025sharp,szymanowicz2025flash3d}. 
% However, their predicted representations are usually most reliable near the input view. 
% When extrapolated to large viewpoint changes, the distribution gap between training and inference viewpoints makes it difficult to represent newly exposed regions, leading to spatial holes, broken geometry, and degraded rendering fidelity, as shown in Fig.~\ref{fig:patchgs_overview}. 
% The second paradigm leverages generative priors, especially video or diffusion based models, to synthesize plausible content for regions outside the observable input view~\cite{ren2025gen3c,yu2024viewcrafter,duan2026navcrafter,cao2025mvgenmaster,kong2025causnvs}. 
% Although such methods can produce visually compelling results, their runtime often scales with the number of generated frames, making full novel view sequence generation expensive.

To mitigate capacity competition while preserving algorithm efficiency, this paper proposes Spackle, a residual learning framework that adaptively injects a small number of augmented 3D Gaussians to enhance single-image NVS. We first use a pretrained 3DGS model to synthesize a target novel view and automatically identify poorly reconstructed disoccluded regions via a binary mask. A compact residual model, replicating the 3DGS architecture, is then trained exclusively on these masked regions. At inference, Gaussians from the pretrained backbone and the residual model are combined for final novel view rendering. This design maintains a fixed total Gaussian count while adaptively allocating additional Gaussians to disoccluded regions, effectively mitigating capacity competition.
This approach maintains a fixed total Gaussian count during training (complying with neural network architectural constraints) while adaptively allocating additional Gaussians to disoccluded regions via the residual branch—effectively mitigating capacity competition.

% To address this issue, we propose Spackle, a lightweight post training framework for extending pretrained feed forward 3D Gaussian models to larger viewpoint changes. 
% Instead of modifying the original Gaussian set, Spackle freezes the pretrained base representation and adaptively introduces auxiliary Gaussians for newly disoccluded regions. 
% These auxiliary Gaussians provide additional capacity exactly where the base model is unreliable, while the original Gaussians continue to preserve the geometry and appearance of visible regions. 
% During training, Spackle learns from video inpainting based pseudo targets generated under large viewpoint changes. 
% At inference time, the video inpainting model is no longer required. 
% Spackle operates together with the frozen base model, identifies unreliable regions using a disocclusion map, and predicts auxiliary Gaussians for completion. 
% This selective extension of the Gaussian set improves large view NVS while preserving the efficiency of the feed forward rendering pipeline.

Our contributions are as follows:
\begin{itemize}
\item We discover and formalize capacity competition: a fixed-Gaussian-budget failure mode in feedforward 3DGS NVS.
\item We propose Spackle: a lightweight residual framework that resolves capacity competition without sacrificing efficiency.
\item We validated Spackle's state-of-the-art performance in large-view single-image NVS via comprehensive experiments.
\end{itemize}

\section{Related Work}

\subsection{Single Image Novel View Synthesis (NVS)}

Early learning-based single-image NVS methods framed the task as direct image-space mapping between source and target views~\cite{zhou2016view,liu2018geometry}, but suffered from limited generalization. To address this, subsequent work introduced stronger geometric priors via intermediate 3D representations, including layered depth images~\cite{shih20203d}, multiplane images (MPIs)~\cite{tucker2020single}, adaptive/tiled MPIs~\cite{han2022single,khan2023tiled}, point-based models~\cite{wiles2020synsin}, and conditional neural radiance fields~\cite{yu2021pixelnerf}. While these advances improved view synthesis quality, they were constrained to local view extrapolation and failed to handle large input-to-target viewpoint deviations.

Recent single-image NVS research has diverged into two complementary paradigms. The first focuses on feedforward reconstruction, which directly predicts renderable 3D representations from image observations~\cite{wang2025vggt,szymanowicz2024splatter,szymanowicz2025flash3d,mescheder2025sharp,rahary2026one}. In particular, feedforward 3D Gaussian Splatting (3DGS) methods enable highly efficient novel-view rendering without per-scene optimization~\cite{szymanowicz2024splatter,szymanowicz2025flash3d,mescheder2025sharp}. However, as noted earlier, fixed-budget 3DGS models struggle to reconstruct disoccluded regions during large viewpoint shifts due to capacity competition.

The second paradigm leverages large-scale generative models (e.g., Stable Diffusion) to inpaint disoccluded regions, including works like ZeroNVS~\cite{sargent2024zeronvs}, CAT3D~\cite{gao2024cat3d}, SplatDiff~\cite{zhang2025high}, Gen3C~\cite{ren2025gen3c}, Bolt3D~\cite{szymanowicz2025bolt3d}, WonderWorld~\cite{yu2025wonderworld}, One2Scene~\cite{wang2026one2scene}, and NavCrafter~\cite{duan2026navcrafter}. These methods use priors learned from massive real-image datasets to generate photorealistic content for disoccluded regions even under extreme viewpoint deviations. However, their reliance on iterative generative sampling incurs prohibitive computational costs, making them unsuitable for low-latency applications.

This paper unifies the strengths of both paradigms, addressing large-view-deviation single-image NVS while maintaining high computational efficiency.

\subsection{Decoupled Frameworks with Generated Supervision}

As mentioned earlier, to simultaneously inpaint disoccluded regions and maintain computational efficiency, many recent NVS and 3D reconstruction methods increasingly adopt a decoupled
strategy that transfers generated supervision or generative priors into 3D representations. Lyra~\cite{bahmani2025lyra} distills the implicit 3D knowledge of video diffusion models into an explicit 3DGS representation for efficient 3D scene reconstruction. Lyra 2.0~\cite{shen2026lyra2} further generates long,
camera-controlled, 3D-consistent video trajectories and uses them to fine-tune feed-forward reconstruction models for high-quality 3D scene recovery. FlashWorld~\cite{li2025flashworld} shifts from generating multi-view images followed by reconstruction to directly producing 3D Gaussian representations,
and improves this process through cross-mode distillation. DiFix3D+~\cite{wu2025difix3d+} uses a single-step diffusion model to enhance rendered pseudo-training views and distill the enhanced signals back into 3D representations. RI3D~\cite{paliwal2025ri3d} uses repair and inpainting diffusion models to
provide pseudo ground truth for 3DGS optimization, separately addressing visible regions and missing regions. SyncFix~\cite{li2026syncfix} enforces cross-view consistency during diffusion-based refinement of reconstructed scenes. These
works show that generated image pairs or generated supervision can provide useful signals for occluded or unseen regions that are difficult to reconstruct from observations alone.

This paper proposes a novel residual-learning framework that mitigates the capacity competition problem in these decoupled methods without sacrificing computational efficiency.

\begin{figure}[!t]
    \centering
    \includegraphics[
        width=0.95\linewidth,
        % height=0.45\textheight,
        keepaspectratio
    ]{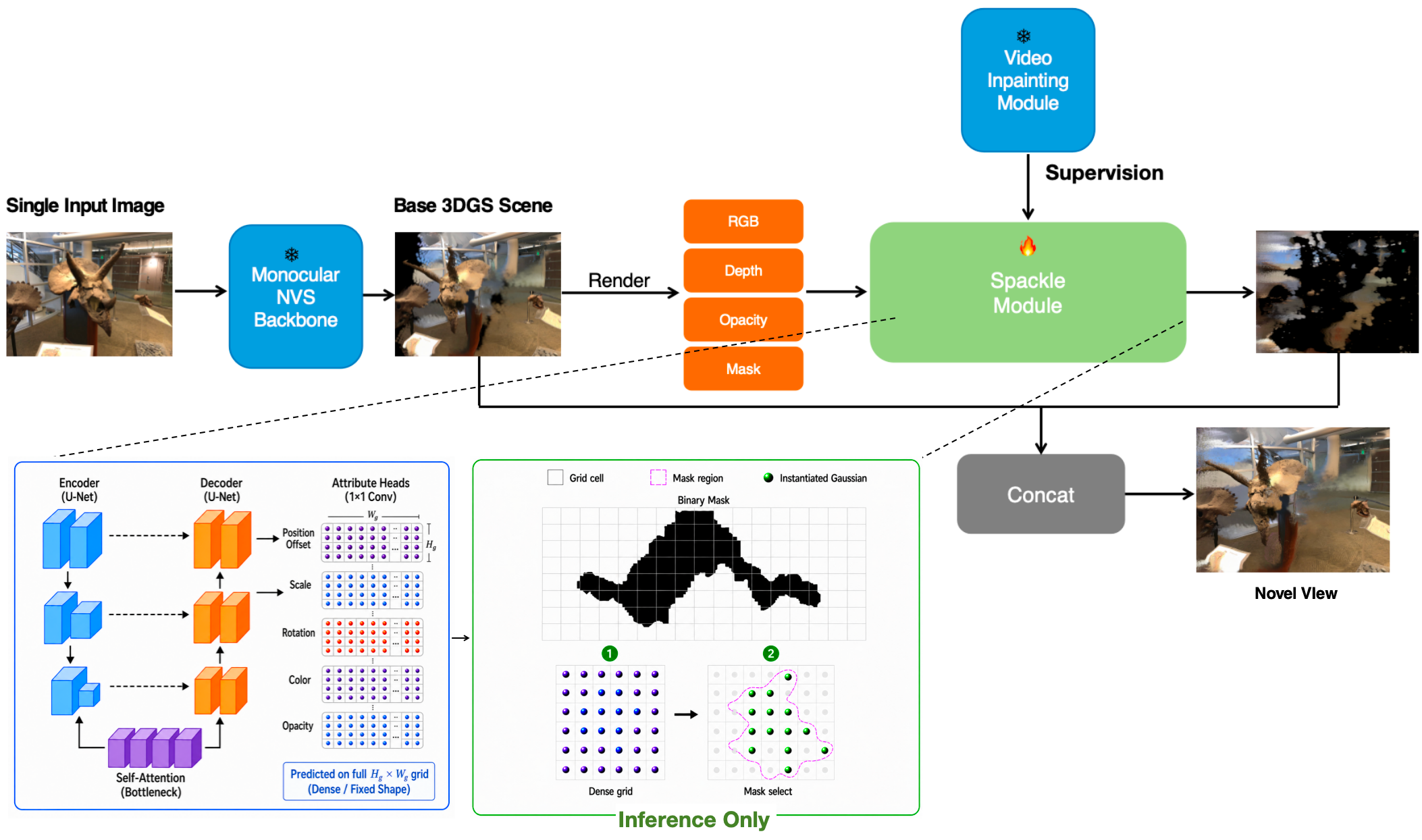}
\caption{
Overview of the Spackle module.
The frozen baseline 3DGS model renders RGB, depth, opacity, and a binary mask for the target novel view.
During training, we use ProPainter~\cite{zhou2023propainter} to generate video inpainting pseudo targets for regions indicated by the binary mask.
Spackle trains a residual 3DGS on these rendered cues to predict augmented Gaussian attributes on a fixed grid.
At inference, only predictions inside the binary mask are converted into augmented Gaussians and combined with the baseline Gaussians for final NVS.
This improves unseen regions that are not well reconstructed while preserving visible-region fidelity.
}
    \label{fig:spackle_module}
\end{figure}

\section{Method}
\subsection{Base 3DGS Prediction}

%Following the decoupled paradigm described above, we first train a baseline feed-forward 3DGS model with a fixed total number of Gaussians across the entire scene. 
Following the decoupled paradigm described above, we adopt a pretrained SHARP~\cite{mescheder2025sharp} model as the baseline 3DGS. Given an input image $I$, the baseline model predicts a 3DGS
scene representation:
\begin{equation}
\mathcal{G}_{\mathrm{base}} = f_{\mathrm{base}}(I),
\label{eq:base_gaussian_prediction}
\end{equation}
where $\mathcal{G}_{\mathrm{base}}$ denotes the baseline Gaussians. The
baseline Gaussians are then rendered from a target viewpoint $v$:
\begin{equation}
\hat{I}_v = \mathcal{R}(\mathcal{G}_{\mathrm{base}}, v).
\label{eq:base_render}
\end{equation}

%In our implementation, we train a SHARP~\cite{mescheder2025sharp} model as the baseline 3DGS model. However, Spackle is not restricted to SHARP and can be applied to different feed-forward 3DGS-based NVS models.
Note that Spackle is not restricted to SHARP~\cite{mescheder2025sharp} and can be applied to other feed-forward 3DGS-based NVS models.
\subsection{Occlusion Identification}

To handle occluded or unobserved regions, we automatically detect such areas in the rendered target view. Given an input image $I$ and a target viewpoint $v$, the baseline 3DGS model predicts Gaussians $\mathcal{G}_{\mathrm{base}}$ and synthesizes the target view:

\begin{equation}
\hat{I}_v, A_v = \mathcal{R}(\mathcal{G}_{\mathrm{base}}, v),
\label{eq:target_novel_view}
\end{equation}

where $\hat{I}_v$ denotes the rendered novel view and $A_v$ represents the corresponding opacity map. Pixels with low opacity indicate occluded or unobserved regions. We formalize these regions via a binary mask:

\begin{equation}
M_v(p) = \mathbbm{1}\left[A_v(p) < \tau_{\alpha}\right],
\label{eq:binary_mask}
\end{equation}

where $p$ denotes a pixel location, $\tau_{\alpha}$ is the opacity threshold, and $M_v(p)=1$ marks occluded or unobserved pixels. Pixels with $M_v(p)=0$ correspond to reliably reconstructed visible areas. This mask subsequently guides the optimization of the residual 3DGS, ensuring that learning is concentrated exclusively on occluded or unobserved regions.

% After training the baseline 3DGS model, we deploy it to synthesize a target novel view. Given an input image $I$ and a target viewpoint $v$, the baseline
% model predicts the baseline Gaussians $\mathcal{G}_{\mathrm{base}}$ and renders
% the target novel view:
% \begin{equation}
% \hat{I}_v, A_v = \mathcal{R}(\mathcal{G}_{\mathrm{base}}, v),
% \label{eq:target_novel_view}
% \end{equation}
% where $\hat{I}_v$ is the rendered target novel view and $A_v$ denotes the
% opacity rendered together with the target novel view. Since the baseline model
% uses a fixed total number of Gaussians across the entire scene, the rendered
% target view may contain unseen regions that are not well reconstructed when the
% target viewpoint deviates significantly from the input camera pose.

% We automatically identify these regions with a binary mask. Specifically, we
% obtain the binary mask from $A_v$:
% \begin{equation}
% M_v(p) = \mathbbm{1}\left[A_v(p) < \tau_{\alpha}\right],
% \label{eq:binary_mask}
% \end{equation}
% where $p$ denotes a pixel location, $\tau_{\alpha}$ is the opacity threshold,
% and $M_v$ denotes the binary mask for the target viewpoint $v$. Pixels with
% $M_v(p)=1$ indicate unseen regions that are not well reconstructed by the
% baseline model, while pixels with $M_v(p)=0$ indicate the remaining regions.
% The resulting binary mask is used in the next step to optimize the residual
% 3DGS exclusively for reconstructing pixels within the mask region.

\subsection{Residual 3DGS Optimization}

Given the binary mask $M_v$ obtained from the occlusion identification step, we train a residual 3DGS model that replicates the baseline architecture but is optimized exclusively for the mask region. Unlike directly optimizing the baseline model, the residual model predicts augmented Gaussians for pixels marked by $M_v$:

\begin{equation}
\mathcal{G}_{\mathrm{aug}} = f_{\mathrm{res}}(I, M_v),
\label{eq:augmented_gaussian_prediction}
\end{equation}

where $f_{\mathrm{res}}$ denotes the residual model and $\mathcal{G}_{\mathrm{aug}}$ represents the augmented Gaussians. This design ensures that the baseline Gaussians $\mathcal{G}_{\mathrm{base}}$ remain unchanged, preserving the fidelity of visible regions.

To train the residual model, we employ a mask-guided reconstruction loss:

\begin{equation}
\mathcal{L}_{\mathrm{mask}}
=
\mathrm{Mean}\Big(
M_v \odot
\left|
\hat{I}_v^{\mathrm{aug}} - I_v^{*}
\right|
\Big),
\label{eq:masked_training_loss}
\end{equation}

where $\hat{I}_v^{\mathrm{aug}}$ denotes the novel view rendered with both baseline and augmented Gaussians, $I_v^{*}$ is the target image, and $\odot$ represents element-wise multiplication. Since the loss is multiplied by $M_v$, only pixels within the mask contribute to optimization, concentrating the learning on occluded or poorly reconstructed regions. 

This strategy preserves the fixed total number of Gaussians during training while effectively enhancing reconstruction in occluded regions, without reallocating capacity from visible areas or modifying the base Gaussians.

\subsection{Residual-Guided Inference}
At inference, Spackle combines the baseline Gaussians with the augmented Gaussians predicted by the residual model to synthesize the final novel view. Given an input image $I$ and a target viewpoint $v$, the baseline 3DGS model predicts $\mathcal{G}_{\mathrm{base}}$, and the residual model predicts augmented Gaussians for the same fixed total number of predictions used during training.

The key difference between training and inference lies in the use of the mask. During inference, only augmented Gaussians corresponding to pixels inside the binary mask $M_v$ are included:

\begin{equation}
\widetilde{\mathcal{G}}_{\mathrm{aug}}(v)
=
\left\{
g_i^{\mathrm{aug}}
\mid
M_v(p_i)=1
\right\},
\label{eq:inference_augmented_gaussians}
\end{equation}

where $p_i$ denotes the location of the $i$-th prediction, and $g_i^{\mathrm{aug}}$ is the corresponding augmented Gaussian. Predictions outside the mask are excluded from $\widetilde{\mathcal{G}}_{\mathrm{aug}}(v)$ and do not participate in rendering.

The final Gaussian set is obtained by combining the baseline and mask-constrained augmented Gaussians:

\begin{equation}
\mathcal{G}_{\mathrm{final}}(v)
=
\mathcal{G}_{\mathrm{base}}
\cup
\widetilde{\mathcal{G}}_{\mathrm{aug}}(v),
\label{eq:final_gaussian_set}
\end{equation}

and the final novel view is rendered as

\begin{equation}
\hat{I}_v^{\mathrm{final}}
=
\mathcal{R}\left(\mathcal{G}_{\mathrm{final}}(v), v\right).
\label{eq:final_rendering}
\end{equation}

This strategy preserves the baseline Gaussians and mitigates capacity competition by restricting the residual model to unseen or poorly reconstructed regions, while maintaining a fixed total Gaussian count during training and inference.

\begin{figure*}[!t]
    \centering
    \includegraphics[
        width=0.95\textwidth,
        % height=0.70\textheight,
        keepaspectratio
    ]{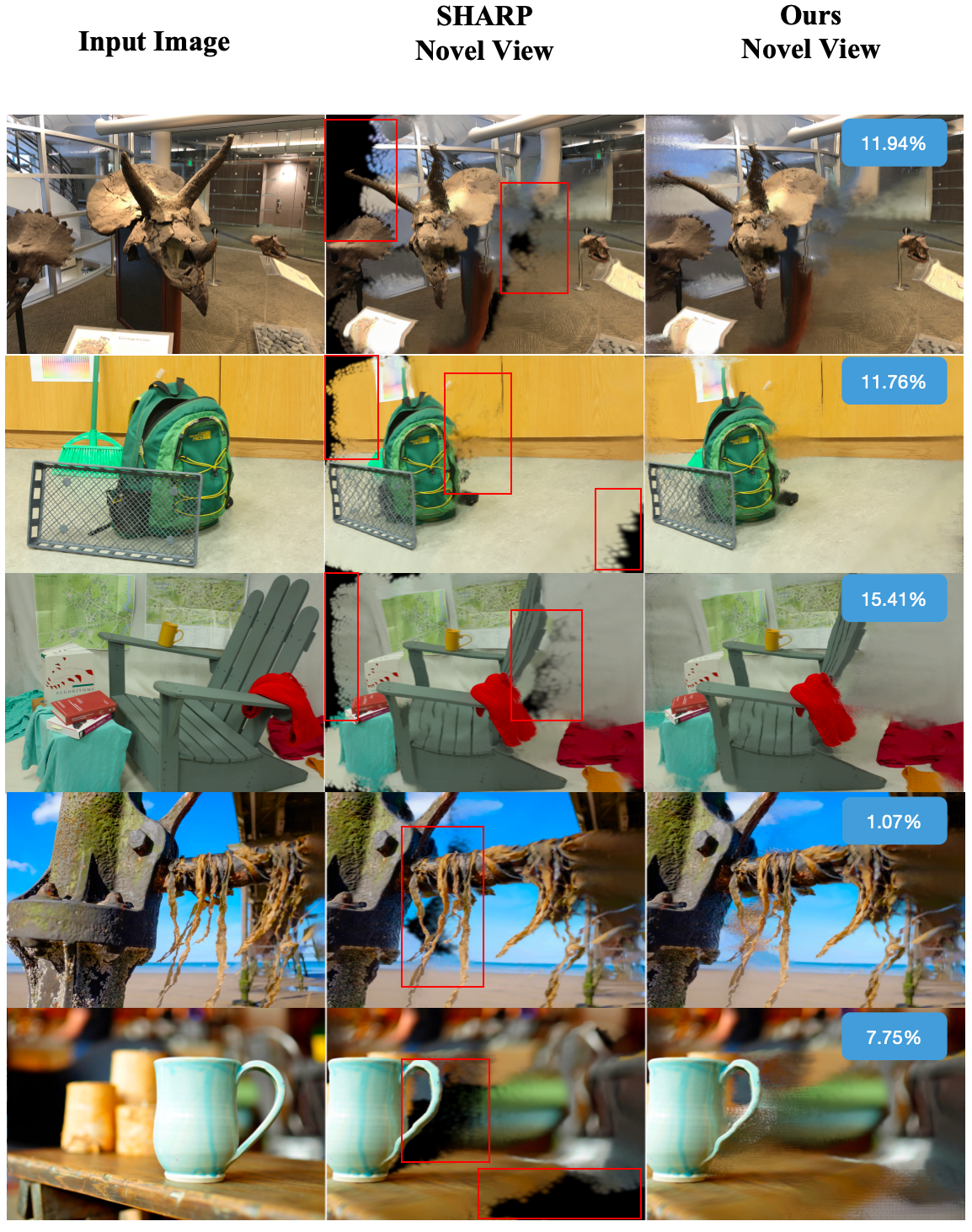}
    \vspace{-10pt}
\caption{
Qualitative comparison under large viewpoint changes.
From left to right, each row shows the input image, the SHARP novel-view rendering with representative artifacts highlighted, and our Spackle novel-view rendering.
The baseline 3DGS model uses a fixed total number of 1,179,648 Gaussians.
The tag at the top-right corner of each Spackle result reports the percentage of augmented Gaussians introduced by Spackle relative to this fixed baseline Gaussian count.
When the target viewpoint deviates significantly from the input camera pose, Spackle better reconstructs unseen regions that are not well reconstructed by the baseline model, while preserving the fidelity of visible regions.
}
    \label{fig:qualitative_results}
\end{figure*}

\section{Experiments}

\subsection{Experimental Setup}

\paragraph{Training Data and Implementation Details.}
Spackle is trained on a subset of the COCO 2017 unlabeled split~\cite{lin2014microsoft}. 
We randomly sample 1,000 images and generate 20 target novel views per image using the frozen baseline 3DGS model, yielding 20,000 training views. 
The rendered target novel views are further completed to provide pseudo supervision for occluded or poorly reconstructed regions. During training, the baseline 3DGS model remains frozen, and only the residual 3DGS is optimized. All renderer-derived inputs are detached before entering Spackle, ensuring that gradients do not modify the baseline Gaussian representation. Training is performed on four NVIDIA RTX 4090 GPUs with 48GB memory each and takes approximately 24 hours.

% We train Spackle on a subset of the COCO 2017 unlabeled split~\cite{lin2014microsoft}. 
% We randomly sample 1,000 images and generate 20 target novel views for each
% image using the frozen baseline 3DGS model, resulting in 20,000 training views.
% The rendered target novel views are further completed to provide pseudo
% supervision for unseen regions that are not well reconstructed. During training,
% the baseline 3DGS model remains frozen, and only the residual 3DGS is optimized.
% All renderer-derived inputs are detached before entering Spackle, ensuring that
% gradients do not modify the baseline Gaussian representation. Training is
% conducted on four NVIDIA RTX 4090 GPUs with 48GB memory each and takes
% approximately 24 hours.

\paragraph{Evaluation Protocol.}
We evaluate Spackle under large viewpoint changes, where the target viewpoint
deviates significantly from the input camera pose. For evaluation, we use images
from Middlebury~\cite{scharstein2014high}, Booster~\cite{ramirez2023booster},
WildRGBD~\cite{xia2024rgbd}, ETH3D~\cite{schops2017multi},
LLFF~\cite{mildenhall2019local}, and Tanks and Temples~\cite{knapitsch2017tanks}.
Following a controlled rendering protocol, we use three increasingly challenging
settings by increasing the disparity from 0.3 to 0.5. Larger disparity exposes
more unseen regions in the target novel view, allowing us to evaluate whether
Spackle can reconstruct these regions while preserving the fidelity of visible
regions. We report qualitative comparisons, no-reference perceptual metrics,
hole coverage, and Gaussian/runtime overhead in the following sections.

\subsection{Main Results}
\paragraph{Qualitative Comparison under Large Viewpoint Deviations.}
We first provide qualitative comparisons under large viewpoint deviations in Fig.~\ref{fig:qualitative_results}, where all visual results are rendered with a challenging disparity of 0.6. When the target viewpoint deviates significantly from the input camera pose, the baseline feed-forward 3DGS model often exhibits visible holes and incomplete structures in newly disoccluded regions. In contrast, Spackle reconstructs these regions more completely while preserving the appearance fidelity of already visible areas. As highlighted by the red boxes, Spackle produces more coherent structures with fewer missing regions under large-view deviations. These results visually demonstrate that learning augmented Gaussians for mask-indicated regions can enhance large-view single-image NVS without disturbing well-reconstructed content. 

We further provide an additional qualitative analysis in Fig.~\ref{fig:direct_ft_failure} in the appendix, where generated supervision is applied directly to the baseline 3DGS model. While this naive approach reduces some missing regions, it often degrades reliably reconstructed visible regions, offering visual evidence of capacity competition.
% We first provide qualitative comparisons under large viewpoint changes in
% Fig.~\ref{fig:qualitative_results}. All visual results are rendered under a
% challenging setting with disparity set to 0.6. When the target viewpoint
% under large viewpoint deviations from the input camera pose, the baseline feed-forward
% 3DGS model often exhibits visible holes and incomplete structures in newly
% disoccluded regions. In contrast, Spackle reconstructs these newly disoccluded regions more
% completely while preserving the appearance fidelity of visible regions. As
% highlighted by the red boxes, Spackle produces more coherent structures and
% fewer missing regions under large-view deviation. These results visually
% demonstrate that learning augmented Gaussians for regions indicated by the
% binary mask can improve large-view single-image NVS without disturbing already
% well-reconstructed content. We further provide an additional qualitative
% analysis of directly applying generated supervision to the baseline 3DGS model
% in Fig.~\ref{fig:direct_ft_failure} in the appendix. The results show that this
% naive decoupled framework may reduce some missing regions but often degrades
% already reliable visible regions, providing visual evidence of capacity
% competition.

\begin{table*}[!t]
\centering
\scriptsize
\setlength{\tabcolsep}{3.4pt}
\renewcommand{\arraystretch}{0.95}
\caption{
Quantitative Evaluation under Controlled Large-View Rendering. Three increasingly challenging novel view settings are generated by increasing disparity. Higher CLIPIQA and MUSIQ scores indicate better perceptual quality, while lower NIQE scores indicate better perceptual quality.
}
\vspace{8pt}
\label{tab:large_view_nr_iqa}
\resizebox{\textwidth}{!}{
\begin{tabular}{llccccccccc}
\toprule
\multirow{2}{*}{Dataset}
& \multirow{2}{*}{Method}
& \multicolumn{3}{c}{S1: Disparity 0.3}
& \multicolumn{3}{c}{S2: Disparity 0.4}
& \multicolumn{3}{c}{S3: Disparity 0.5} \\
\cmidrule(lr){3-5}
\cmidrule(lr){6-8}
\cmidrule(lr){9-11}
&
& CLIPIQA $\uparrow$
& NIQE $\downarrow$
& MUSIQ $\uparrow$
& CLIPIQA $\uparrow$
& NIQE $\downarrow$
& MUSIQ $\uparrow$
& CLIPIQA $\uparrow$
& NIQE $\downarrow$
& MUSIQ $\uparrow$ \\
\midrule
\multirow{2}{*}{Middlebury}
& SHARP   & 0.3904 & 4.3102 & 58.2782 & 0.3913 & 4.2697 & 57.7797 & 0.3926 & 4.1992 & 56.8124 \\
& Spackle & \best{0.3921} & 4.3127 & 58.1583 & \best{0.3969} & 4.2825 & 57.6487 & \best{0.4032} & 4.2268 & \best{56.9233} \\
\midrule
\multirow{2}{*}{Booster}
& SHARP   & 0.4221 & 5.1467 & 55.7906 & 0.4314 & 5.1630 & 55.0354 & 0.4397 & 5.1894 & 54.6178 \\
& Spackle & \best{0.4224} & \best{5.1466} & \best{55.7929} & \best{0.4325} & \best{5.1625} & 54.9989 & \best{0.4443} & \best{5.1868} & 54.5166 \\
\midrule
\multirow{2}{*}{WildRGBD}
& SHARP   & 0.2782 & 4.3972 & 45.2073 & 0.2595 & 4.4207 & 42.3572 & 0.2451 & 4.4680 & 39.9257 \\
& Spackle & \best{0.2858} & \best{4.3422} & \best{45.2259} & \best{0.2676} & \best{4.2775} & \best{42.7294} & \best{0.2533} & \best{4.2471} & \best{40.6128} \\
\midrule
\multirow{2}{*}{Tanks}
& SHARP   & 0.3536 & 3.1289 & 60.9697 & 0.3666 & 3.2049 & 60.2709 & 0.3820 & 3.3647 & 59.8877 \\
& Spackle & \best{0.3793} & 3.1858 & \best{61.1473} & \best{0.3955} & 3.2137 & \best{60.4960} & \best{0.4126} & \best{3.2472} & \best{60.1183} \\
\midrule
\multirow{2}{*}{ETH3D}
& SHARP   & 0.3820 & 3.5688 & 64.7636 & 0.3863 & 3.5976 & 63.5485 & 0.3994 & 3.6497 & 62.5826 \\
& Spackle & \best{0.4020} & 3.5732 & \best{65.0792} & \best{0.4126} & 3.6095 & \best{63.9991} & \best{0.4274} & \best{3.6353} & \best{63.1081} \\
\midrule
\multirow{2}{*}{LLFF}
& SHARP   & 0.4601 & 3.6783 & 65.2599 & 0.4574 & 3.6185 & 64.1105 & 0.4566 & 3.6132 & 63.0390 \\
& Spackle & \best{0.4719} & \best{3.6670} & \best{65.4440} & \best{0.4779} & 3.6251 & \best{64.2562} & \best{0.4843} & \best{3.6028} & \best{63.1472} \\
\bottomrule
\end{tabular}
}
\end{table*}

\begin{table*}[!t]
\centering
\scriptsize
\setlength{\tabcolsep}{4.2pt}
\renewcommand{\arraystretch}{1.2}
\caption{
Large view hole coverage evaluation on benchmark datasets.
Controlled camera changes with increasing disparity expose newly disoccluded regions.
We report Hole Area Ratio (HAR), defined as $1-\mathrm{Coverage}$ from the rendered opacity map.
Lower HAR indicates fewer remaining holes, and Reduction measures the relative HAR decrease of Spackle over SHARP.
}
\vspace{8pt}
\label{tab:controlled_disocclusion}
\resizebox{\textwidth}{!}{
\begin{tabular}{lccccccccc}
\toprule
\multirow{2}{*}{Dataset}
& \multicolumn{3}{c}{S1: Disparity 0.3}
& \multicolumn{3}{c}{S2: Disparity 0.4}
& \multicolumn{3}{c}{S3: Disparity 0.5} \\
\cmidrule(lr){2-4}
\cmidrule(lr){5-7}
\cmidrule(lr){8-10}
& \tiny SHARP HAR $\downarrow$
& \tiny Spackle HAR $\downarrow$
& \tiny Reduction $\uparrow$
& \tiny SHARP HAR $\downarrow$
& \tiny Spackle HAR $\downarrow$
& \tiny Reduction $\uparrow$
& \tiny SHARP HAR $\downarrow$
& \tiny Spackle HAR $\downarrow$
& \tiny Reduction $\uparrow$ \\
\midrule
Middlebury
& 0.0121 & 0.0011 & 90.47\%
& 0.0305 & 0.0035 & 88.52\%
& 0.0566 & 0.0084 & 85.13\% \\

Booster
& 0.0036 & 0.0003 & 91.84\%
& 0.0103 & 0.0006 & 94.51\%
& 0.0242 & 0.0011 & 95.40\% \\

WildRGBD
& 0.0503 & 0.0043 & 91.37\%
& 0.0885 & 0.0166 & 81.29\%
& 0.1286 & 0.0338 & 73.75\% \\

Tanks
& 0.1464 & 0.0369 & 74.77\%
& 0.2177 & 0.0569 & 73.83\%
& 0.2794 & 0.0620 & 77.81\% \\

ETH3D
& 0.0680 & 0.0120 & 82.33\%
& 0.1181 & 0.0261 & 77.87\%
& 0.1720 & 0.0406 & 76.41\% \\

LLFF
& 0.0440 & 0.0045 & 89.67\%
& 0.0818 & 0.0129 & 84.18\%
& 0.1226 & 0.0229 & 81.29\% \\
\bottomrule
\end{tabular}
}
\end{table*}

\paragraph{Quantitative Evaluation under Large Viewpoint Deviations.}
We further assess perceptual quality for large-view single-image NVS using CLIPIQA~\cite{wang2023exploring}, NIQE~\cite{mittal2012making}, and MUSIQ~\cite{ke2021musiq}. Following the evaluation protocol in Sec.~4.1, results are reported under three increasingly challenging disparity settings from 0.3 to 0.5. As shown in Tab.~\ref{tab:large_view_nr_iqa}, Spackle consistently improves CLIPIQA scores across most datasets and viewpoint settings, with more pronounced gains under larger viewpoint deviations. This indicates that Spackle produces more plausible and semantically coherent novel views when substantial occluded regions are exposed.

NIQE and MUSIQ show smaller or mixed changes compared with CLIPIQA, which is expected since these generic no-reference metrics are not specifically tailored for evaluating localized reconstruction of unseen regions. Nevertheless, Spackle maintains competitive perceptual quality while significantly enhancing the visual completeness of large-view renderings. Together with qualitative results, these observations demonstrate the effectiveness of Spackle for single-image NVS under large viewpoint deviations.
% \paragraph{Large-View No-Reference Perceptual Evaluation.}
% We further evaluate perceptual quality under large viewpoint changes using
% CLIPIQA~\cite{wang2023exploring}, NIQE~\cite{mittal2012making}, and
% MUSIQ~\cite{ke2021musiq}. Following the evaluation protocol in Sec.~4.1, we
% report results under three increasingly challenging disparity settings from
% 0.3 to 0.5. As shown in Tab.~\ref{tab:large_view_nr_iqa}, Spackle consistently
% improves CLIPIQA across most datasets and viewpoint settings, with more visible
% gains under larger viewpoint deviations. This indicates that Spackle produces
% more plausible and semantically coherent novel views when substantial unseen
% regions are exposed.

% NIQE and MUSIQ show smaller or mixed changes compared with CLIPIQA. This is
% reasonable because these generic no-reference metrics are not specifically
% designed for evaluating localized reconstruction of unseen regions. Nevertheless,
% Spackle maintains competitive perceptual quality while improving the visual
% completeness of large-view renderings. Together with the qualitative results,
% these observations support the effectiveness of Spackle for large-view-deviation
% single-image NVS.

\paragraph{Large-View Unseen-Region Coverage Evaluation.}
We also evaluate whether Spackle reduces remaining uncovered regions in rendered
novel views. Different from no-reference perceptual metrics, this evaluation
directly measures the uncovered regions using the rendered opacity map. Given
the opacity map $\alpha_v$ of a rendered target view, we compute the coverage
ratio as
\begin{equation}
\mathrm{Coverage}
=
\frac{1}{HW}
\sum_{p}
\mathbf{1}\left[\alpha_v(p)\ge\tau_\alpha\right],
\end{equation}
and define the Hole Area Ratio as
\begin{equation}
\mathrm{HAR}
=
1-\mathrm{Coverage}.
\end{equation}
Lower HAR indicates fewer uncovered pixels. As reported in
Tab.~\ref{tab:controlled_disocclusion}, Spackle consistently
reduces HAR across six datasets and three disparity settings. This confirms that the
augmented Gaussians effectively improve coverage in unseen regions that are not
well reconstructed under large-view extrapolation.

\subsection{Ablation Studies}

\paragraph{Naive Decoupled Framework vs. Residual Learning.}
We first compare Spackle with a naive decoupled framework to verify the effect
of residual learning. In the naive decoupled framework, generated supervision is
directly injected into the baseline 3DGS model. Please refer to
Tab.~\ref{tab:near_view} in the appendix for the quantitative results. This
baseline consistently degrades DISTS and LPIPS across datasets, supporting our
observation that directly applying generated supervision to the baseline 3DGS
model with a fixed total number of Gaussians can harm already well-reconstructed
visible regions. In contrast, Spackle preserves the baseline 3DGS model and
trains a residual 3DGS only for the binary mask indicating unseen regions that
are not well reconstructed. This allows Spackle to improve large-view
reconstruction without degrading standard view rendering quality.

\paragraph{Binary-Mask-Guided Gaussian Allocation.}
We further ablate the binary-mask-guided Gaussian allocation strategy. Please
refer to Tab.~\ref{tab:mask_guided_init} in the appendix for the quantitative
results. Compared with dense grid initialization, binary-mask-guided allocation
only introduces augmented Gaussians for the mask region. This greatly reduces
the number of augmented Gaussians while maintaining nearly identical
DISTS~\cite{ding2020iqa} and LPIPS~\cite{zhang2018unreasonable}. These
results show that Spackle does not need to introduce redundant Gaussians over
the entire image. Instead, the binary mask provides an effective constraint for
allocating additional Gaussians only to unseen regions that are not well
reconstructed.

\subsection{Efficiency and Gaussian Analysis}

\paragraph{Gaussian and Runtime Overhead.}
We analyze the Gaussian and runtime overhead of Spackle in
Tab.~\ref{tab:efficiency_gaussian_overhead}. The baseline SHARP model uses a
fixed set of 1,179,648 Gaussians across the entire scene. In contrast, Spackle
introduces augmented Gaussians only for the binary mask indicating unseen
regions that are not well reconstructed. As the disparity increases from 0.3 to
0.5, the average number of augmented Gaussians increases from 11.5K to 28.0K,
showing that Spackle adaptively allocates additional Gaussians according to the
size of newly disoccluded regions. The median runtime overhead remains moderate,
increasing from 2.7\% at S1 to 4.4\% at S3. These results show that Spackle
improves large-view reconstruction with limited additional Gaussian and runtime
cost.

\paragraph{Expanded Efficiency Analysis.}
We further provide an expanded efficiency analysis in
Tab.~\ref{tab:expanded_efficiency_gaussian_overhead} in the appendix. This table
reports additional statistics including parameter overhead, rendering time, and
FPS under controlled large-view settings. SHARP has 702.31M parameters, while
Spackle has 710.32M parameters, introducing only 8.02M additional parameters,
corresponding to 1.14\% overhead. These results further confirm that Spackle
mitigates capacity competition and improves large-view NVS while preserving the
feed-forward efficiency required for practical deployment.

\begin{table*}[t]
\centering
\scriptsize
\setlength{\tabcolsep}{4.2pt}
\renewcommand{\arraystretch}{0.95}
\caption{
Efficiency and Gaussian overhead under controlled large view settings.
SHARP uses a fixed base set of 1,179,648 Gaussians. 
Spackle adaptively introduces additional Gaussians according to the disocclusion regions.
Runtime is reported as the median over datasets for each setting to reduce the influence of measurement outliers.
}
\vspace{10pt}
\label{tab:efficiency_gaussian_overhead}
\resizebox{\textwidth}{!}{
\begin{tabular}{lcccccc}
\toprule
\multirow{2}{*}{Dataset}
& \multicolumn{2}{c}{S1: Disparity 0.3}
& \multicolumn{2}{c}{S2: Disparity 0.4}
& \multicolumn{2}{c}{S3: Disparity 0.5} \\
\cmidrule(lr){2-3}
\cmidrule(lr){4-5}
\cmidrule(lr){6-7}
& Added GS & Added GS (\%)
& Added GS & Added GS (\%)
& Added GS & Added GS (\%) \\
\midrule
Middlebury & 3,387  & 0.29 & 8,501  & 0.72 & 15,782 & 1.34 \\
Booster    & 1,074  & 0.09 & 3,077  & 0.26 & 7,226  & 0.61 \\
WildRGBD   & 3,823  & 0.32 & 6,755  & 0.57 & 9,826  & 0.83 \\
Tanks      & 33,817 & 2.87 & 50,254 & 4.26 & 64,466 & 5.46 \\
ETH3D      & 18,663 & 1.58 & 32,382 & 2.75 & 47,105 & 3.99 \\
LLFF       & 8,425  & 0.71 & 15,629 & 1.32 & 23,397 & 1.98 \\
\midrule
Average    & 11,532 & 0.98 & 19,433 & 1.65 & 27,967 & 2.37 \\
\midrule
\multicolumn{7}{c}{Runtime efficiency} \\
\midrule
Median SHARP Time (ms)   & \multicolumn{2}{c}{1412.0} & \multicolumn{2}{c}{1411.8} & \multicolumn{2}{c}{1411.8} \\
Median Spackle Time (ms) & \multicolumn{2}{c}{1449.8} & \multicolumn{2}{c}{1454.4} & \multicolumn{2}{c}{1474.0} \\
Median Time Overhead     & \multicolumn{2}{c}{2.7\%}  & \multicolumn{2}{c}{3.0\%}  & \multicolumn{2}{c}{4.4\%} \\
\bottomrule
\end{tabular}
}
\end{table*}

\FloatBarrier

\section{Conclusion and Limitations}

We presented Spackle, a residual learning framework for large-view single-image novel view synthesis built upon a pretrained feed-forward 3DGS backbone. Spackle mitigates capacity competition while preserving efficient feed-forward rendering. Our key observation is that fixed-budget 3DGS models struggle under large viewpoint deviations because the same Gaussian set must represent both visible and newly disoccluded regions. Spackle addresses this through mask-guided adaptive residual Gaussian augmentation, allocating additional Gaussian capacity exclusively to occluded regions via sparse instantiation. During training, generative completion priors are distilled into the residual branch to provide pseudo supervision, while inference remains purely feed-forward without requiring diffusion or video inpainting models. Experiments demonstrate that Spackle consistently improves large-view completion while preserving the fidelity and efficiency of feed-forward 3DGS rendering.
% We presented Spackle, a residual learning framework for large-view single-image
% novel view synthesis that mitigates capacity competition while preserving
% efficient feed-forward rendering. Our key observation is that fixed-budget
% feed-forward 3DGS models struggle under large viewpoint changes because the same
% Gaussian set must represent both visible and newly disoccluded regions.
% Spackle addresses this issue through adaptive residual Gaussian augmentation,
% allocating additional Gaussian capacity only to disoccluded regions via adaptive
% sparse instantiation. During training, generative completion priors are distilled
% into the residual branch, while inference remains purely feed-forward without
% requiring diffusion or video inpainting models. Experiments demonstrate that
% Spackle consistently improves large-view completion while preserving the fidelity
% and efficiency of feed-forward 3DGS rendering.

Spackle’s performance still relies on accurate disocclusion estimation and the quality of pseudo supervision. Additionally, residual learning introduces modest computational overhead, and results may degrade if the baseline 3DGS model is poorly initialized. Future work may explore more robust disocclusion estimation, improved pseudo-supervision strategies, and scalable residual Gaussian allocation for challenging large-view scenarios.
% Spackle still depends on the quality of disocclusion estimation and pseudo
% supervision, and extremely large viewpoint extrapolation remains inherently
% ambiguous in the single-image setting. Future work may explore more robust
% disocclusion estimation and temporally consistent residual Gaussian augmentation
% for long-range camera trajectories.

\small
\bibliographystyle{plain}
\bibliography{references}
\newpage
\appendix

\section*{Technical appendices and supplementary material}
\section{Additional Quantitative Ablations}

We provide additional quantitative ablations to further analyze the behavior of
Spackle. First, Tab.~\ref{tab:near_view} compares SHARP, the naive decoupled
framework, and Spackle under the same evaluation setting. In the naive
decoupled framework, generated supervision is directly injected into the
baseline 3DGS model. This consistently worsens DISTS and LPIPS across datasets,
suggesting that applying generated supervision to a baseline 3DGS model with a
fixed total number of Gaussians can degrade already well-reconstructed visible
regions. In contrast, Spackle preserves the baseline 3DGS model and achieves
comparable or better results than SHARP in most cases, showing that the
augmented Gaussians do not noticeably harm standard view rendering quality.

Second, Tab.~\ref{tab:mask_guided_init} evaluates binary-mask-guided Gaussian
allocation. Compared with dense grid initialization, the proposed strategy
greatly reduces the number of augmented Gaussians while maintaining nearly
identical DISTS and LPIPS. This indicates that Spackle can adaptively allocate
additional Gaussians only to unseen regions that are not well reconstructed,
rather than introducing augmented Gaussians over the entire grid.

\begin{table}[htbp]
\centering
\scriptsize
\setlength{\tabcolsep}{2.0pt}
\renewcommand{\arraystretch}{0.95}
\caption{
Quantitative comparison between SHARP, the naive decoupled framework, and Spackle. Lower is better.
The naive decoupled framework directly optimizes the baseline 3DGS model with pseudo supervision, while Spackle preserves the baseline 3DGS model and trains augmented Gaussians for newly disoccluded regions.
}
\label{tab:near_view}
\begin{tabular}{lcccccccccccccc}
\toprule
& \multicolumn{2}{c}{Middlebury}
& \multicolumn{2}{c}{Booster}
& \multicolumn{2}{c}{WildRGBD}
& \multicolumn{2}{c}{Tanks}
& \multicolumn{2}{c}{ETH3D}
& \multicolumn{2}{c}{LLFF} \\
\cmidrule(lr){2-3}
\cmidrule(lr){4-5}
\cmidrule(lr){6-7}
\cmidrule(lr){8-9}
\cmidrule(lr){10-11}
\cmidrule(lr){12-13}
\cmidrule(lr){14-15}
Method
& DISTS & LPIPS
& DISTS & LPIPS
& DISTS & LPIPS
& DISTS & LPIPS
& DISTS & LPIPS
& DISTS & LPIPS \\
\midrule
SHARP
& 0.0875 & 0.3675
& 0.0872 & 0.3076
& 0.0451 & 0.2002
& 0.5238 & 0.7392
& 0.3644 & 0.6627
& 0.4837 & 0.7154 \\

Direct FT
& 0.1506 & 0.4447
& 0.1390 & 0.3740
& 0.0662 & 0.2414
& 0.5698 & 0.7602
& 0.4501 & 0.7261
& 0.6042 & 0.7325 \\

Spackle
& 0.0874 & 0.3674
& 0.0871 & 0.3075
& 0.0451 & 0.2004
& 0.4839 & 0.7313
& 0.3318 & 0.6425
& 0.4589 & 0.7180 \\
\bottomrule
\end{tabular}
\end{table}

\begin{table}[htbp]
\centering
\scriptsize
\setlength{\tabcolsep}{4.5pt}
\renewcommand{\arraystretch}{0.95}
\caption{
Ablation on disocclusion guided sparse initialization across evaluation datasets.
Compared with dense grid initialization, disocclusion guided initialization reduces the number of initialized auxiliary Gaussians while maintaining rendering quality.
Base GS denotes the fixed number of Gaussians predicted by SHARP, and Aux. GS denotes the average number of initialized auxiliary Gaussians per novel view.
}
\label{tab:mask_guided_init}
\begin{tabular}{llccccc}
\toprule
Dataset
& Method
& Base GS
& Aux. GS $\downarrow$
& Reduction $\uparrow$
& DISTS $\downarrow$
& LPIPS $\downarrow$ \\
\midrule
Middlebury
& Dense grid init
& \multirow{14}{*}{1,179,648}
& 414,018.78
& --
& 0.0874 & 0.3674 \\
& Mask guided init
&
& 330.65
& 99.92\%
& 0.0874 & 0.3674 \\
\cmidrule(lr){1-2}\cmidrule(lr){4-7}

Booster
& Dense grid init
&
& 431,616.00
& --
& 0.0871 & 0.3075 \\
& Mask guided init
&
& 144.17
& 99.97\%
& 0.0871 & 0.3075 \\
\cmidrule(lr){1-2}\cmidrule(lr){4-7}

WildRGBD
& Dense grid init
&
& 76,800.00
& --
& 0.0450 & 0.2002 \\
& Mask guided init
&
& 91.11
& 99.88\%
& 0.0451 & 0.2004 \\
\cmidrule(lr){1-2}\cmidrule(lr){4-7}

Tanks
& Dense grid init
&
& 331,776.00
& --
& 0.4839 & 0.7313 \\
& Mask guided init
&
& 193,131.43
& 41.79\%
& 0.4839 & 0.7313 \\
\cmidrule(lr){1-2}\cmidrule(lr){4-7}

ETH3D
& Dense grid init
&
& 393,246.72
& --
& 0.3319 & 0.6425 \\
& Mask guided init
&
& 167,031.80
& 57.52\%
& 0.3318 & 0.6425 \\
\cmidrule(lr){1-2}\cmidrule(lr){4-7}

LLFF
& Dense grid init
&
& 190512.00
& --
& 0.4589 & 0.7180 \\
& Mask guided init
&
& 96824.65
& 49.18\%
& 0.4589 & 0.7180\\
\bottomrule
\end{tabular}
\end{table}

\begin{table*}[htbp]
\centering
\scriptsize
\setlength{\tabcolsep}{3.2pt}
\renewcommand{\arraystretch}{0.95}
\caption{
Expanded efficiency and Gaussian overhead analysis under controlled large view settings.
SHARP uses a fixed base set of 1,179,648 Gaussians.
Spackle adds auxiliary Gaussians according to the disocclusion regions.
The parameter overhead is constant across settings: SHARP has 702.31M parameters, while Spackle has 710.32M parameters, introducing 8.02M additional parameters, corresponding to 1.14\%.
Time is measured in milliseconds, and FPS denotes rendering throughput.
A negative time overhead indicates that Spackle is faster than SHARP in the corresponding aggregated entry, which can occur due to runtime measurement outliers.
}
\label{tab:expanded_efficiency_gaussian_overhead}
\resizebox{\textwidth}{!}{
\begin{tabular}{llrrrrrrrrr}
\toprule
Dataset 
& Setting 
& Added GS 
& Added GS (\%) 
& Merged GS 
& SHARP Time 
& Spackle Time 
& Time Overhead 
& SHARP FPS 
& Spackle FPS 
& $n$ \\
\midrule
\multirow{3}{*}{Middlebury}
& S1 & 3,387  & 0.29 & 1,183,035 & 6410.3 & 1449.4 & -3.3\% & 0.671 & 0.691 & 23 \\
& S2 & 8,501  & 0.72 & 1,188,149 & 1448.8 & 1452.7 & 0.8\%  & 0.695 & 0.689 & 23 \\
& S3 & 15,782 & 1.34 & 1,195,430 & 1448.4 & 1476.4 & 2.5\%  & 0.695 & 0.679 & 23 \\
\midrule
\multirow{3}{*}{Booster}
& S1 & 1,074 & 0.09 & 1,180,722 & 1427.8 & 1451.4 & 1.9\% & 0.703 & 0.691 & 228 \\
& S2 & 3,077 & 0.26 & 1,182,725 & 1427.6 & 1456.1 & 2.3\% & 0.703 & 0.689 & 228 \\
& S3 & 7,226 & 0.61 & 1,186,874 & 1427.9 & 1471.6 & 3.3\% & 0.703 & 0.682 & 228 \\
\midrule
\multirow{3}{*}{WildRGBD}
& S1 & 3,823 & 0.32 & 1,183,471 & 1419.8 & 1443.8 & 1.9\% & 0.709 & 0.697 & 512 \\
& S2 & 6,755 & 0.57 & 1,186,403 & 1419.6 & 1449.3 & 2.3\% & 0.709 & 0.694 & 512 \\
& S3 & 9,826 & 0.83 & 1,189,474 & 1419.6 & 1459.5 & 3.0\% & 0.709 & 0.689 & 512 \\
\midrule
\multirow{3}{*}{Tanks}
& S1 & 33,817 & 2.87 & 1,213,465 & 1404.1 & 1519.5 & 8.4\%  & 0.716 & 0.661 & 512 \\
& S2 & 50,254 & 4.26 & 1,229,902 & 1403.9 & 1566.8 & 11.8\% & 0.716 & 0.641 & 512 \\
& S3 & 64,466 & 5.46 & 1,244,114 & 1403.9 & 1607.5 & 14.7\% & 0.716 & 0.625 & 512 \\
\midrule
\multirow{3}{*}{ETH3D}
& S1 & 18,663 & 1.58 & 1,198,311 & 1371.2 & 1450.1 & 5.8\%  & 0.730 & 0.691 & 454 \\
& S2 & 32,382 & 2.75 & 1,212,030 & 1371.0 & 1490.4 & 8.8\%  & 0.730 & 0.673 & 454 \\
& S3 & 47,105 & 3.99 & 1,226,753 & 1370.9 & 1533.4 & 11.9\% & 0.730 & 0.654 & 454 \\
\midrule
\multirow{3}{*}{LLFF}
& S1 & 8,425  & 0.71 & 1,188,073 & 1360.8 & 1405.3 & 3.3\% & 0.735 & 0.712 & 305 \\
& S2 & 15,629 & 1.32 & 1,195,277 & 1360.6 & 1424.1 & 4.7\% & 0.735 & 0.703 & 305 \\
& S3 & 23,397 & 1.98 & 1,203,045 & 1360.5 & 1447.3 & 6.4\% & 0.735 & 0.692 & 305 \\
\bottomrule
\end{tabular}
}
\end{table*}

\section{Additional Qualitative Analysis of the Naive Decoupled Framework}

\begin{figure*}[!t]
    \centering
    \includegraphics[
        width=\textwidth,
        height=0.72\textheight,
        keepaspectratio
    ]{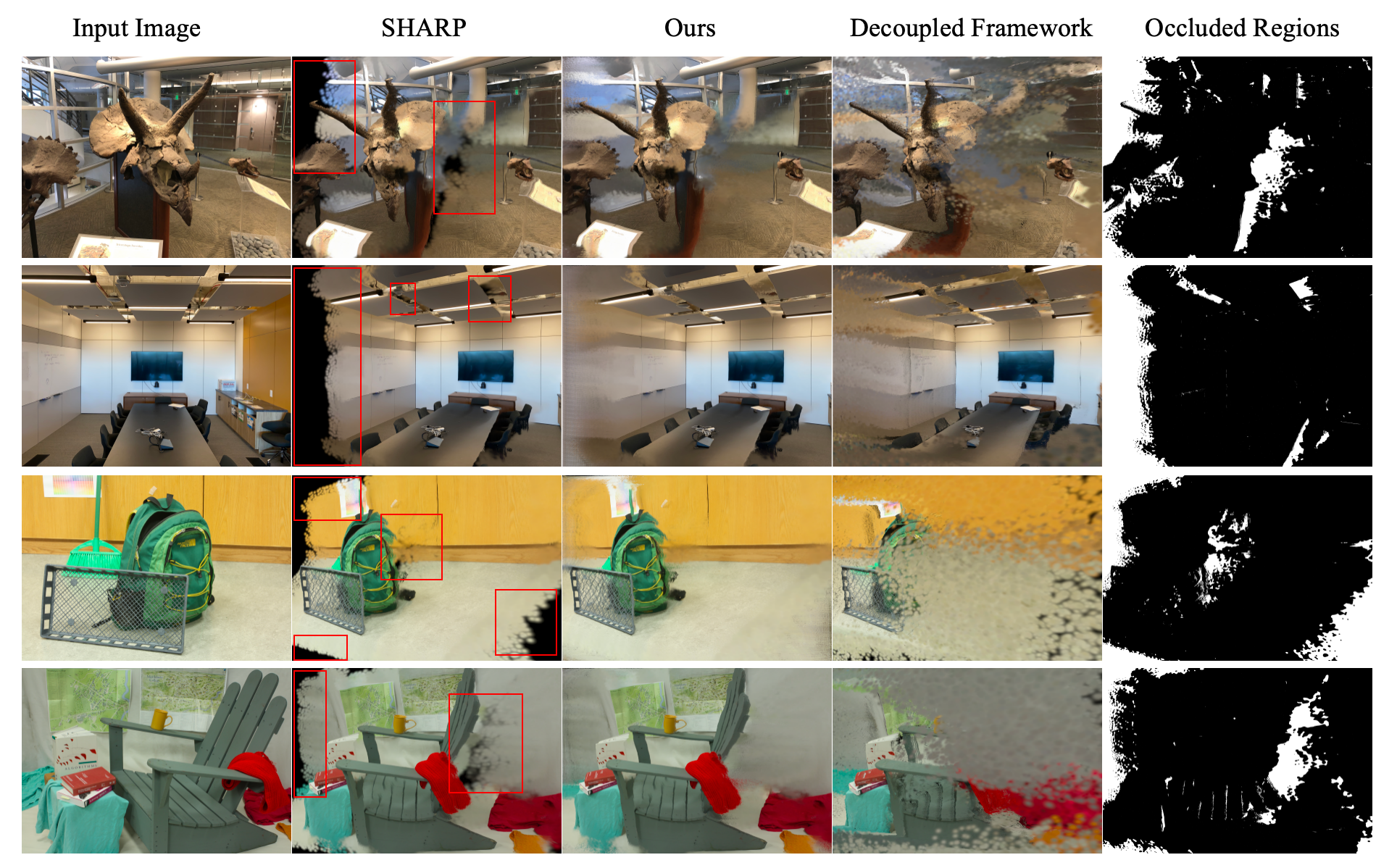}
\caption{
Additional qualitative analysis of directly applying generated supervision to
the baseline 3DGS model under large viewpoint changes. From left to right, each
row shows the input image, the SHARP rendering, our Spackle rendering, the
rendering from the naive decoupled framework, and the detected occluded regions.
The naive decoupled framework can reduce some missing regions, but it often
introduces noticeable appearance degradation in already reliable areas. In
contrast, Spackle keeps the baseline 3DGS model frozen and introduces augmented
Gaussians for newly disoccluded regions, preserving the main scene appearance
while improving hole completion.
}
    \label{fig:direct_ft_failure}
\end{figure*}

Fig.~\ref{fig:direct_ft_failure} provides additional qualitative evidence for
the capacity competition phenomenon discussed in the main paper. Directly
applying generated supervision to the baseline 3DGS model can improve some
newly disoccluded regions, but it may also degrade already well-reconstructed
visible regions. In contrast, Spackle preserves the baseline 3DGS model and
learns a residual 3DGS optimized exclusively for the binary mask indicating
unseen regions that are not well reconstructed.
%%%%%%%%%%%%%%%%%%%%%%%%%%%%%%%%%%%%%%%%%%%%%%%%%%%%%%%%%%%%

\section{Expanded Efficiency and Gaussian Overhead Analysis}

Tab.~\ref{tab:expanded_efficiency_gaussian_overhead} provides a detailed efficiency breakdown for each dataset and controlled large view setting. 
Compared with the fixed SHARP base set of 1,179,648 Gaussians, Spackle adds only a small fraction of auxiliary Gaussians, and the added amount increases with disparity as more disoccluded regions are exposed. 
The table also reports merged Gaussians, runtime, FPS, and the number of evaluated samples. 
Overall, Spackle maintains moderate runtime overhead while adaptively allocating additional Gaussian capacity for large view completion. 
The parameter overhead is constant across settings because the same Spackle module is used for all views.

\section{Pseudo Supervised Dataset Construction}

We construct the pseudo supervised training set from the COCO 2017 unlabeled split~\cite{lin2014microsoft}. 
We randomly sample 1,000 images and use the frozen feed forward reconstructor to predict the final base Gaussian representation for each image. 
The base Gaussians are rendered along a controlled camera trajectory with 20 views, including the original input view and rendered novel views, producing 20,000 training views in total.

For each sampled image, we save the input image, rendered RGB frames, rendered opacity maps, rendered depth maps, and per-frame camera metadata. 
The camera metadata records the trajectory type, camera positions, intrinsics, extrinsics, and image resolution for each rendered frame, so every pseudo target can be paired with its exact target viewpoint. 
The disocclusion masks are derived from the rendered opacity maps: pixels with opacity below a threshold are treated as uncovered regions, where 255 indicates an uncovered pixel and 0 indicates a covered pixel.

We then apply ProPainter~\cite{zhou2023propainter} to each rendered trajectory using the corresponding mask sequence. 
We choose ProPainter because it is not diffusion based, making the inpainting process more controllable and efficient than iterative diffusion based generation. 
It can complete masked regions over a video sequence with temporal consistency, while requiring relatively modest computation to process a large number of rendered trajectories. 
These properties make it suitable for constructing pseudo targets for Spackle, where the video inpainting model is used only during dataset construction rather than at inference time. 
The inpainted frames are used as pseudo target images for supervising Spackle in newly disoccluded regions.

Each final training sample contains the input image, target camera parameters, base rendering, opacity, depth, disocclusion mask, and completed pseudo target. 
The video inpainting model is used only for constructing the pseudo supervised dataset, while inference only uses the frozen base reconstructor and Spackle.

\section{Training Details and Loss Weights}

We train Spackle on the pseudo supervised dataset described above while keeping the pretrained base reconstructor frozen. 
Only the Spackle branch is optimized, and all parameters of the base feed forward reconstructor remain fixed. 
We use a learning rate of $1.6\times 10^{-4}$ with a cosine decay schedule after linear warmup. 
The minimum learning rate is set to $1.6\times 10^{-5}$, and the warmup length is 10,000 steps. 
Early stopping is enabled with patience 10 and a minimum improvement threshold of $10^{-4}$. 
During training, the hole mask is predicted from the frozen base rendering opacity using an alpha threshold of 0.95. 
We also keep at least 64 initialized auxiliary Gaussians per novel view to avoid degenerate empty predictions.

The training objective contains four active terms. 
Let $\tilde{I}_v$ and $\tilde{\alpha}_v$ denote the rendered RGB image and opacity map after adding auxiliary Gaussians for target view $v$, and let $I_v^{*}$ denote the video inpainting pseudo target. 
Let $Q_v^{\mathrm{dis}}$ denote the disocclusion mask, where $Q_v^{\mathrm{dis}}(p)=1$ indicates that pixel $p$ belongs to a hole region. 
The color loss is an L1 loss between $\tilde{I}_v$ and $I_v^{*}$ on the hole region. 
The alpha loss applies binary cross entropy to encourage $\tilde{\alpha}_v$ to approach one inside the hole region. 
The perceptual loss uses a fixed ResNet-50 feature extractor and combines feature reconstruction with Gram statistics; the feature term is computed on a dilated hole context using a dilation kernel of 21 and feature resolution 512. 
Finally, the non-hole preservation loss matches the Spackle rendering to the detached no-Spackle rendering outside the hole region, preventing auxiliary Gaussians from changing regions already explained by the base model.

The final objective is
\begin{equation}
\mathcal{L}
=
10.0\,\mathcal{L}_{\mathrm{color}}
+
0.5\,\mathcal{L}_{\alpha}
+
1.0\,\mathcal{L}_{\mathrm{perc}}
+
5.0\,\mathcal{L}_{\mathrm{preserve}} .
\end{equation}
The input-view losses in the implementation are used only for the non-Spackle training path and are disabled in the main Spackle setting. 
All reported Spackle results are trained with the above objective.

\section{Broader Impacts}

Spackle may benefit applications such as virtual and augmented reality, 3D content creation, and embodied intelligence by improving large-view novel view synthesis from a single image. 
However, because Spackle completes regions that are not directly observed in the input image, it may also be misused to generate plausible but incorrect 3D scene content or misleading visual evidence. 
We therefore view Spackle as a research tool for novel view synthesis rather than a source of verified scene evidence.

\section{Existing Assets and Licenses}

We use existing public datasets, pretrained models, and evaluation metrics in this work. 
The training images are sampled from the COCO 2017 unlabeled split~\cite{lin2014microsoft}. 
We use SHARP~\cite{mescheder2025sharp} as the frozen base reconstructor and ProPainter~\cite{zhou2023propainter} to generate video inpainting pseudo targets. 
For evaluation, we use images from Middlebury~\cite{scharstein2014high}, Booster~\cite{ramirez2023booster}, WildRGBD~\cite{xia2024rgbd}, ETH3D~\cite{schops2017multi}, LLFF~\cite{mildenhall2019local}, and Tanks and Temples~\cite{knapitsch2017tanks}. 
We also use standard no-reference image quality metrics, including CLIPIQA~\cite{wang2023exploring}, NIQE~\cite{mittal2012making}, and MUSIQ~\cite{ke2021musiq}. 
All assets are used for academic research purposes, and we follow the licenses and terms of use provided by their original authors.

\end{document}